\documentclass[10pt, a4paper]{article}
\usepackage{lrec}
\usepackage{multibib}
\newcites{languageresource}{Language Resources}
\usepackage{graphicx}
\usepackage{tabularx}
\usepackage{soul}

\usepackage{booktabs}

\usepackage{xcolor}
\usepackage{tikz}
\usepackage{pgfplots, pgfplotstable}
\usetikzlibrary{patterns}

\usepackage{epstopdf}
\usepackage[latin1]{inputenc}
\usepackage{authblk}
\usepackage{hyperref}
\usepackage{xstring}
\usepackage{amsmath}

\newcommand{\COMMENT}[1]{}

\title{Examining the Tip of the Iceberg: \\ A Data Set for Idiom Translation}

\name{Marzieh Fadaee\textsuperscript{1}, Arianna Bisazza\textsuperscript{2}, Christof Monz\textsuperscript{1}}

\address{\textsuperscript{1}Informatics Institute, University of Amsterdam,
Science Park 904, 1098 XH Amsterdam, The Netherlands\\
\textsuperscript{2}Leiden Institute of Advanced Computer Science,
Leiden University,
2333 CA Leiden,
The Netherlands\\
\texttt{\{m.fadaee,c.monz\}@uva.nl}\\ \texttt{a.bisazza@liacs.leidenuniv.nl}}

\abstract{
Neural Machine Translation (NMT) has been widely used in recent years with significant improvements for many language pairs. 
Although state-of-the-art NMT systems are generating progressively better translations, idiom translation remains one of the open challenges in this field. 
Idioms, a category of multiword expressions, are an interesting language phenomenon where the overall meaning of the expression cannot be composed from the meanings of its parts.
A first important challenge is the lack of dedicated data sets for learning and evaluating idiom translation.
In this paper we address this problem by creating the first large-scale data set for idiom translation. 
Our data set is automatically extracted from a widely used German$\leftrightarrow$English translation corpus and includes, for each language direction, a targeted evaluation set where all sentences contain idioms and a regular training corpus where sentences including idioms are marked.
We release this data set and use it to perform preliminary NMT experiments as the first step towards better idiom translation. 
\\
\newline  }

\begin{document}

\maketitleabstract

\section{Introduction}

Neural Machine Translation (NMT) \cite{DBLP:journals/corr/BahdanauCB14,sutskever2014sequence,cho2014properties} has achieved substantial improvements in translation quality over traditional Rule-based and Phrase-based Translation (PBMT) in recent years. 
For instance, subject-verb agreement, double-object verbs, and overlapping subcategorization are various areas where NMT successfully overcomes the limitations of PBMT \cite{isabelle2017challenge,bentivogli-EtAl:2016:EMNLP2016}.
However, one of the remaining challenges of NMT is translating infrequent words and phrases \cite{koehn2017six,fadaee-bisazza-monz:2017:Short2} and idioms are a particular instance of this problem \cite{isabelle2017challenge}. 

Idioms are semantic lexical units whose meaning is often not simply a function of the meaning of its constituent parts \cite{10.2307/416483,doi:10.1093/applin/17.3.326}.
The non-compositionality characteristic of idiom expressions exists in different degrees in a language \cite{10.2307/416483}.
In English for example, for the idiom \textit{``spill the beans"}, the word \textit{`spill'} symbolizes \textit{`reveal'} and \textit{`beans'} symbolizes the \textit{`secrets'}. 
With the idiomatic expression \textit{``kick the bucket"}, on the other hand, no such analysis is possible.

\begin{table}[htb!]
\centering
\small
\begin{tabularx}{\linewidth}{p{2.7cm} X}
 \toprule
German phrase & \textit{eine wei{\ss}e Weste haben}  \\
 Literal translation & to have a white vest \\
 Idiomatic translation  &  to have clean slate  \\
	\midrule
 Sentence & Coca-Cola und Nestl{\'e} geh{\"o}ren zu den Unterzeichnern. Beide \textbf{haben} nicht gerade \textbf{eine wei{\ss}e Weste}. \\
  Reference translation & Coca Cola and Nestl{\'e} are two signatories with \textit{``spotty" track records}.\\
  \midrule
 DeepL   & Coca-Cola and Nestl{\'e} are among the signatories. Neither of them is \textbf{exactly the same}.  \\
GoogleNMT  & Coca-Cola and Nestl{\'e} are among the signatories. Both do not \textbf{have just a white vest}.  \\
OpenNMT & Coca-Cola and Nestl{\'e} are among the signatories. Both don't \textbf{have a white essence}.  \\
\bottomrule
\end{tabularx}
\caption{Example of an idiom phrase in German and its translation. We compare the output of DeepL, GoogleNMT, and OpenNMT translating a sentence with this idiom phrase and notice that none capture the idiom translation correctly.}
\label{examplesinc}
\end{table}

\newcite{isabelle2017challenge} builds a challenge set of 108 short sentences that each focus on one particular difficult phenomenon of the language.
Their manual assessment of the eight sentences consisting of an idiomatic phrase show that NMT systems struggle with the translation of these phrases.

The challenge of translating idiom phrases in NMT is partly due
to the underlying complexity of identifying a phrase as idiomatic and generating its correct non-literal translation, and partly
to the fact that idioms are rarely encountered in the standard data sets used for training NMT systems. 

As an example, in Table~\ref{examplesinc} we provide an idiom expression in German and the literal and idiomatic translations in English.
We observe that the literal translation of an idiom is not the correct translation, neither does it capture part of the meaning. 

\begin{figure*}[ht!]
\centering
\includegraphics[width=0.95\linewidth]{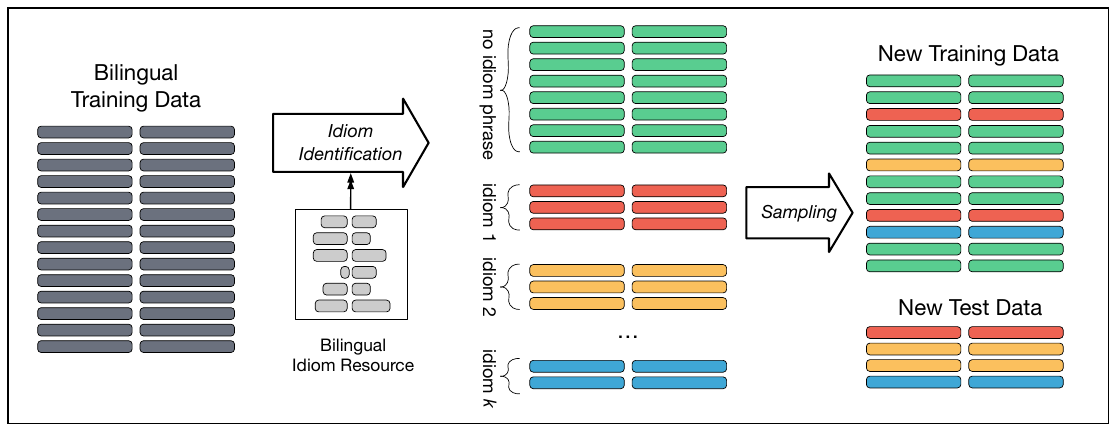}
\caption{The process of data collection and construction of the test set containing only sentence pairs with idiom phrases.}
\label{augfig}
\end{figure*}

To illustrate the problem of idiom translation we also provide the output of three 
NMT systems for this sentence: GoogleNMT \cite{wu2016google}, DeepL\footnote{\url{www.deepl.com/translator}}, and the OpenNMT implementation \cite{2017opennmt} based on \newcite{DBLP:journals/corr/BahdanauCB14} and \newcite{luong:2015:EMNLP}. 
All systems fail to generate the proper translation of the idiom expression. 
This problem is particularly pronounced when the source idiom is very different from its equivalent in the target language, as the case here.

Although there are a number of monolingual data sets available for identifying idiom expressions \cite{muzny2013automatic,markantonatou2017proceedings}, there is limited work on building a parallel corpus annotated with idioms, which is necessary to investigate this problem more systematically.  
\newcite{salton-ross-kelleher:2014:HyTra} selected a small subset of 17 English idioms, collected 10 sentence examples for each idiom from the internet, and manually translated them into Brazilian-Portuguese to use for the translation task.

Building a hand-crafted data set for idiom translation is costly and time-consuming.
In this paper we automatically build a new bilingual data set for idiom translation 
extracted from an existing general-purpose German$\leftrightarrow$English parallel corpus.

The first part of our data set consists of 1,500 parallel sentences whose German side contains an idiom, while the second consists of 1,500 parallel sentences whose English side contains an idiom.
%
Additionally, we provide the corresponding training data sets for German$\rightarrow$English and English$\rightarrow$German translation where source sentences including an idiom phrase are marked. 
We believe that having a sizable data set for training and evaluation is the first step to improve idiom translation.


 \begin{table}[htb!]
\centering
\small
\begin{tabular}{l r}
 \toprule
  German idiom translation data set & \\
  \midrule
Number of unique idioms & 103   \\
  Training size & 4.5M \\
  Idiomatic sentences in training data & 1848 \\
 Test size & 1500  \\
\toprule
  English idiom translation data set & \\
  \midrule
Number of unique idioms &  132  \\
  Training size & 4.5M \\
  Idiomatic sentences in training data & 1998  \\
 Test size & 1500  \\
\bottomrule
\end{tabular}
\caption{Statistics of the German and English idiom translation data sets. Sentence pairs are counted on the training and test sets.}
\label{stats}
\end{table}

\section{Data Collection} \label{data}

In this work we focus on German$\leftrightarrow$English translation of idioms. 
This is an established language pair and is commonly used in the machine translation community.

Automatically identifying idiom phrases in a parallel corpus requires a gold standard data set annotated manually by linguists. 
We use the \url{dict.cc} online dictionary\footnote{\texttt{www.dict.cc}} containing idiomatic and colloquial phrases, which is built manually, as our gold standard for extracting idiom phrase pairs. 

Examining the WMT German$\leftrightarrow$English test sets from 2008 to 2016 \cite{bojar-EtAl:2017:WMT1}, we observe very few sentence pairs containing an idiomatic expression. 
The standard parallel corpora available for training however contain several such sentence pairs.
Therefore we automatically select sentence pairs from the training corpora where the source sentence contains an idiom phrase to build the new test set.

Note that we only focus on idioms on the source side and we have two separate list of idioms for German and English, hence, we independently build two test sets (for German idiom translation and English idiom translation) with different sentence pairs selected from the parallel corpora.

\begin{table}[htb!]
\centering
\small
\begin{tabularx}{\linewidth}{@{\ }l @{\ \ \ }X@{\ }}
 \toprule
German idiom & \textit{alles {\"u}ber einen kamm scheren}  \\
English equivalent & to measure everything by the same yardstick \\
Matching German sentence & Aber man kann eben nicht \textbf{alle} Inseln \textbf{\"{u}ber einen Kamm scheren}.\\
English translation &  But we cannot measure everyone by the same standards.\\ 
 \midrule
German idiom & \textit{in den kinderschuhen stecken}  \\
English equivalent & to be in the fledgling stage \\
Matching German sentence & Es \textbf{steckt} immer noch \textbf{in den Kinderschuhen}. \\
English translation &  It is still in its infancy. \\
\bottomrule
\end{tabularx}
\caption{Two examples displaying different constraints of matching an idiom phrase with occurrences in the sentence.}
\label{two}
\end{table}

  \begin{table*}[ht]
 \begin{tabularx}{\linewidth}{lX}
       \toprule
       German idiom & \textit{in den kinderschuhen stecken} \\
       English equivalent &  to be in the fledgling stage   \\
       German sentence &     Eine Bemerkung, Gentoo/FreeBSD \textbf{steckt} noch \textbf{in den Kinderschuhen} und ist kein auf Sicherheit achtendes System. \\
       English sentence &     Note that Gentoo/FreeBSD is still \textbf{in its infancy} and is not a security supported platform. \\
       \midrule
       German idiom & \textit{den kreis schlie{\ss}en}   \\
       English equivalent  &  to bring sth. full circle      \\
       German sentence  & Die europ{\"a}ische Krise \textbf{schlie{\ss}t den Kreis}.\\
       English sentence  & The European crisis is \textbf{coming full circle}. \\
\midrule
       German idiom &    \textit{auf biegen und brechen} \\
       English equivalent   & by hook or crook    \\
       German sentence  & Nehmen wir zum Beispiel die W{\"a}hrungsunion: Sie soll \textbf{auf Biegen und Brechen} eingef{\"u}hrt werden.\\
       English sentence  & Take, for example, the introduction -\textbf{come what may}- of the single currency.\\
\midrule
       German idiom &  \textit{sie haben das wort} \\
       English equivalent  &   the floor is yours    \\
       German sentence  & Berichterstatterin. - (FR) Herr Pr{\"a}sident! Danke, dass \textbf{Sie} mir \textbf{das Wort} erteilt \textbf{haben}.\\
       English sentence  & rapporteur. - (FR) Mr President, thank you for \textbf{giving} me \textbf{the floor}.\\
\bottomrule
 \end{tabularx}
 \caption{Examples from the German idiom translation test set.}
 \label{biggest}
 \end{table*}

Depending on the language, the words making up an idiomatic phrase are not always contiguous in the sentence. 
For instance, in German, the subject can appear between the verb and the prepositional phrase making up the idiom. 
German also allows for several re-orderings of the phrase.

In order to generalize the process of identifying idiom occurrences, we lemmatize the phrases and consider different re-ordering of the words in the phrase as an acceptable match. 
We also allow for a fixed number of words to occur in between the words of an idiomatic phrase.
Table~\ref{two} shows two examples of idiom occurrences that match these criteria.

Following this set of rules, we extract sentence pairs containing idiomatic phrases, and create a set of sentence pairs for each unique idiom phrase.
In the next step we sample without replacement from these sets and select individual sentence pairs to build the test set.

In order to build the new training data, we use the remaining sentence pairs in each idiom set as well as the sentence pairs from the original parallel corpora that did not include any idiom phrases.
In this process, we ensure that for each idiomatic expression there is at least one occurrence in both training and test data, and that no sentence is included in both training and test data. 

Figure~\ref{augfig} visualizes the process of constructing the new training and test sets.
As a result, for each language direction, we obtain a targeted test set for idiom translation and the corresponding training corpus representing a natural distribution of sentences with and without idioms.


We annotate each sentence pair with the canonical form of its source-side idiom phrase and its equivalent in the target language. 

Table~\ref{stats} provides some statistics of the two data sets. 
For each unique idiom in the test set, we also provide the frequency of the respective idiom in the training data. 
Note that this is based on the lemmatized idiom phrase under the constraints mentioned in Section~\ref{data} and is not necessarily an exact match of the phrase.


Table~\ref{biggest} shows several examples from the data set for German idiom translation. 
We observe that for some idioms the literal translation in the target language is close to the actual meaning, while for others it is not the case.


\begin{table*}[ht!]
\begin{center}
\begin{tabular}{r | c | c c c}
 & \multicolumn{1}{c}{WMT test sets 2008-2016}  & \multicolumn{3}{c}{Idiom test set}\\
      \toprule       
     Model &  BLEU  &  BLEU  &   Unigram Precision & Word-level Accuracy  \\
\midrule
PBMT Baseline & 20.2  & 19.7  &  57.7 & 71.6 \\
NMT Baseline   &  26.9  &    24.8       &  53.2 &   67.8  \\
NMT \texttt{<idm>} token on source & 25.2  &  22.5       & 64.1 & 73.2     \\
      \bottomrule
\end{tabular}
\caption{Translation performance on German idiom translation test set. \textit{Word-level Idiom Accuracy} and \textit{Unigram Precision} are computed only on the idiom phrase and its corresponding translation in the sentence.}
  \label{numbers}
 \end{center}
\end{table*}

One side effect of automatically identifying idiom expressions in sentences is that it is not always accurate. 
Sentence pairs where an idiom expression was used as a literal phrase (e.g., \textit{``spill the beans"}  to literally describe the act of \textit{spilling the beans}) will be identified as idiomatic sentences.   

\section{Translation Experiments}
    
While the main focus of this work is to generate data sets for training and evaluating idiom translation, we also perform a number of preliminary NMT experiments using our data set to measure the problem of idiom translation on large scale data. 

In the first experiment following the conventional settings, we do not use any labels in the data to train the translation model.
In the second experiment we use the labels in the training data as an additional feature to investigate the effect of informing the model of the existence of an idiomatic phrase in a sentence during training.

We perform 
a German$\rightarrow$English experiment by providing the model with additional input features.
The additional features indicate whether a source sentence contains an idiom
and are implemented as a special extra token \texttt{<idm>} that is prepended to each source sentence containing an idiom.
This a simple approach that can be applied to any sequence-to-sequence architecture. 

Most NMT systems have a sequence-to-sequence architecture where an encoder builds up a representation of the source sentence and a decoder, using the previous LSTM hidden states and an attention mechanism, generates the target translation \cite{DBLP:journals/corr/BahdanauCB14,sutskever2014sequence,cho2014properties}.
We use a 4-layer attention-based encoder-decoder model as described in \cite{luong:2015:EMNLP} trained with hidden dimension size of 1,000, and batch size of 80 for 20 epochs. 

In all experiments the NMT vocabulary is limited to the most common 30K words in both languages and we preprocess source and target language data with Byte-pair encoding (BPE) \cite{sennrich-haddow-birch:2016:P16-12} using 30K merge operations. 

We also use a Phrase-based translation system similar to Moses \cite{koehn2007moses} as baseline to explore PBMT performance for idiom translation.

\section{Idiom Translation Evaluation}

Ideally idiom translation should be evaluated manually, but this is a very costly process.
Automatic metrics, on the other hand, can be used on large data sets at no cost and have the advantage of replicability.

We use the following metrics to evaluate the translation quality with a specific focus on idiom translation accuracy:

\paragraph{BLEU} The traditional BLEU score \cite{Papineni2001} is a good measure to determine the overall quality of the translations. 
However this measure considers the precision of all $n$-grams  in a sentence and by itself does not focus on the translation quality of the idiomatic expressions.

\paragraph{Modified Unigram Precision} 
To specifically concentrate on the quality of the translation of idiom expressions, we also look at the \textit{localized} precision. 
In this approach we translate the idiomatic expression in the context of a sentence, and only evaluate the translation quality of the idiom phrase. 

To isolate the idiom translation in the sentence, we look at the word-level alignments between the idiom expression in the source sentence and the generated translation in the target sentence. 
We use \texttt{fast-align} \cite{dyer-chahuneau-smith:2013:NAACL-HLT} to extract word alignments.
Since idiomatic phrases and the respective translations are not contiguous in many cases we only compare the unigrams of the two phrases.


Note that for this metric 
we have two references: The idiom translation as an independent expression, and the human generated idiom translation in the target sentence. 

\paragraph{Word-level Idiom Accuracy} We also use another metric to evaluate the word-level translation accuracy of the idiom phrase. 
We use word alignments between source and target sentences to determine the number of correctly translated words.
We use the following equation to compute the accuracy:
\begin{align*} 
WIAcc = \frac{H-I}{N}
\end{align*} 

where $H$ is the number of correctly translated words, $I$ is the number of extra words in the idiom translation, and $N$ is the number of words in the gold idiom expression.

\smallskip


Table~\ref{numbers} presents the results for the translation task using different metrics. 
Looking at the overall BLEU scores, we observe that baseline performance on the idiom-specific test set is lower than on the union of the standard test sets (WMT 2008-2016). While the scores on these two data sets are not directly comparable,
this result is in line with previous findings that sentences containing idiomatic expressions are harder to translate \cite{isabelle2017challenge}.
We can also see that the performance gap is 
not as pronounced for PBMT systems, suggesting that phrase-based models are capable of \textit{memorizing} the idiom phrases to some extent.

The NMT experiment using a special input token to indicate the presence of an idiom in the sentence performs still better than PBMT but slightly worse than the NMT baseline in terms of BLEU.
Despite this drop in BLEU performance, by examining the \textit{unigram precision} and \textit{word-level idiom accuracy} scores, we observe that this model generates more accurate idiom translations.

These preliminary experiments reiterate the problem of idiom translation with neural models, and in addition show that with a labeled data set, we can devise simple models to address this problem to some extent.




\section{Conclusion}
Idiom translation is one of the more difficult challenges of machine translation. Neural MT in particular has been shown to perform poorly on idiom translation despite its overall strong advantage over previous MT paradigms \cite{isabelle2017challenge}. 
As a first step towards a better understanding of this problem,
we have presented a parallel data set for training and testing idiom translation for German$\rightarrow$English and English$\rightarrow$German.

The test sets include sentences with at least one idiom on the source side while the training data is a mixture of idiomatic and non-idiomatic sentences with labels to distinguish between the two.
We also performed preliminary translation experiments and proposed different metrics to evaluate idiom translation.

We release new data sets which can be used to further investigate and improve NMT performance in idiom translation.

\section*{Acknowledgments}
This research was funded in part by the Netherlands Organization for Scientific Research (NWO) under project numbers 639.022.213 and 639.021.646, and a Google Faculty Research Award. We also thank NVIDIA for their hardware support. 

\section{Bibliographical References}
\label{main:ref}

\bibliographystyle{lrec}
\bibliography{lrec}


\end{document}